\documentclass[journal]{IEEEtran}
\usepackage{graphicx}
\usepackage{epstopdf}
\usepackage{gensymb}
\usepackage[table]{xcolor}
\usepackage{caption}
\usepackage[noadjust]{cite}
\usepackage{physics}

\ifCLASSINFOpdf
\else
\fi

\begin{document}
\bstctlcite{IEEEexample:BSTcontrol}

\title{DeepFeat: A Bottom Up and Top Down Saliency Model Based on Deep Features of Convolutional Neural Nets}

\author{Ali~Mahdi,~\IEEEmembership{Student Member,~IEEE,}
        and~Jun~Qin,~\IEEEmembership{Member,~IEEE}
\thanks{A. Mahdi and J. Qin are with the ECE Department, Southern Illinois Univ., Carbondale, IL, 62901 USA e-mail: (ali.mahdi@siu.edu; jqin@siu.edu).}
}

\markboth{Submitted manuscript to IEEE TCDS}%
{Shell \MakeLowercase{\textit{et al.}}: Bare Demo of IEEEtran.cls for IEEE Journals}
%

\maketitle

\begin{abstract}

A deep feature based saliency model (DeepFeat) is developed to leverage the understanding of the prediction of human fixations. Traditional saliency models often predict the human visual attention relying on few level image cues. Although such models predict fixations on a variety of image complexities, their approaches are limited to the incorporated features. In this study, we aim to provide an intuitive interpretation of convolutional neural network deep features by combining low and high level visual factors. We exploit four evaluation metrics to evaluate the correspondence between the proposed framework and the ground-truth fixations. The key findings of the results demonstrate that the DeepFeat algorithm, incorporation of bottom up and top down saliency maps, outperforms the individual bottom up and top down approach. Moreover, in comparison to nine 9 state-of-the-art saliency models, our proposed DeepFeat model achieves satisfactory performance based on all four evaluation metrics.

\end{abstract}


\IEEEpeerreviewmaketitle

\section{Introduction}

\IEEEPARstart{T}{he} human visual system has an exceptional ability of sampling the surrounding world to pay attention to objects of interest. Such ability is the visual attention that guides the visual exploration. Visual attention requires a complex cognitive mechanism to allocate the human gaze toward the objects of interest. In computer vision, a saliency map is defined to model the human visual attention. A saliency map is a 2D topological map that indicates visual attention priorities in a numerical scale. A higher visual attention priority indicates the object of interest is irregular or rare to its surroundings. The modeling of saliency is beneficial for several applications including image segmentation \cite{mishra2009active, maki2000attentional}, object detection \cite{butko2009optimal,ehinger2009modelling}, image re-targeting \cite{marchesotti2009framework,suh2003automatic}, image/video compression \cite{itti2004automatic, guo2010novel}, and advertising design \cite{rubinstein2008improved}, etc.

The research on saliency modeling is influenced by bottom up and top down visual cues. The bottom up visual attention (exogeneous) is triggered by stimulus, where a saliency is captured as the distinction of image locations, regions, or objects in terms of low level cues such as color, intensity, orientation, shape, T-conjunctions, X-conjunctions, etc \cite{nothdurft2005salience}. One of the bottlenecks bottom up saliency models suffer, is that they explain the scene partially as majority of the human eye fixations are task driven. Following the feature integration theory (FIT) \cite{treisman1980feature}, the first saliency model was proposed \cite{itti1998model}. The model exploit the biologically inspired center-surround scheme of color, intensity, and orientation at various scales to identify distinctive image locations. Bruce \& Tsotsos proposed an attentional information maximization model to predict eye fixations \cite{bruce2005saliency}. The model uses self information to detect saliency in local image regions. Zhang et al. derived a Bayesian framework that incorporates self information of local image regions with prior knowledge about the image \cite{zhang2008sun}. Liu t al. developed a saliency model as a decision tree of regional saliency measurements including global contrast, spatial sparsity, and object prior \cite{liu2014saliency}. Zhang \& Sclaroff developed a saliency map based on a boolean approach. The model combine birnary maps and attention maps \cite{zhang2016exploiting}. The binary maps are obtained via random thresholding of the color feature of the image. Attention maps are computed using the gestalt principle of the figure-ground segregation. Leboran et al. proposed a dynamic whitening saliency model to predict fixations in videos. The model uses whitening to access the relevant information by removing the second order information.

The top down visual attention is driven by task. Top down saliency models use prior knowledge, expectations, or rewards as high level visual factors to identify the target of interest \cite{borji2013state}. Several top down saliency models have been proposed. Such as, Oliva et al. introduced a top down visual search model based on Bayesian framework. The model exploits cognitive features and scales \cite{oliva2003top}. Contextual features are represented by reducing dimensionality of local features. The joint probability of a feature vector is computed using multivariate Gaussian distributions. Rao proposed an attention representation as a cortical mechanism for reducing perceptual uncertainty. The model exploits belief propagation in a probabilistic framework to combine bottom up and top down visual factors \cite{rao2005bayesian}. Judd et al. developed a saliency model to predict where human look by combining low, mid, and high level cues as visual features \cite{judd2009learning}, and used support vector machines to learn to predict human fixations. Borji et al. proposed a saliency model based on top down factors to learn task driven object based visual attention control in interacting environment \cite{borji2010online}. Recently, Wang et al. combined 13 bottom up and top down saliency models using several combination strategies \cite{wang2016learning}. Then trained the model using support vector machine.

Recently, deep features of the deep neural networks (DNN) have been used in several applications, inculding imaging and video processing, medical signal processing, big data analysis, and saliency modeling as well \cite{sun2016neural, zhao2015saliency, qin2015engineering, han2015background, sun2016enhanced}. Although the intuition of the DNN deep features remain unclear \cite{chu2017visualizing, zeiler2014visualizing, yosinski2015understanding}, several saliency models used pre-trained deep features to detect bottom up and top down visual cues. Deep features are the response images of convolution, batch normalization, activation, and pooling operations in a series of layers in a convolutional neural network \cite{schmidhuber2015deep}. Such response images provide semantic information about the image. Initial layers present low level cues such as edges, and a higher level abstract is obtained as a function of layer number. Latter layers provide higher level of semantic information such as a class of objects. 

Although intensive research effort intended to leverage the understanding of human visual attention \cite{mahdi2017comparison, mahdi2015infants, le2017visual}, classical saliency models suffer a few bottlenecks such as feature selection. The task of selecting features to integrate in a saliency model is overwhelming, because the saliency model identifies salient locations in terms of the pre-defined features. To overcome this bottleneck, we introduce a framework to combine pre-trained deep features of a convolutional neural network. The proposed framework defines the deep features as bottom up and top down visual cues. The rest of this section provides a literature review of deep learning saliency models that utilize the pre-trained deep features, and our contribution in this study.

\subsection{Related Work}
The recent research efforts aim to impact saliency prediction using deep learning models such as convolutional nerual network (CNN), recurrent neural network (RNN), or deep belief network (DBN), etc. In image processing, CNN is ideal because in local image patches pixels correlate to each other. CNN based saliency models exploit state-of-the-art CNNs such as AlexNet \cite{krizhevsky2012imagenet}, VGG \cite{simonyan2014very}, GoogleNet \cite{szegedy2015going}, and ResNet \cite{he2016deep}. The first saliency model based on deep learning is proposed by Vig et al. \cite{vig2014large}. The architucture of the model consists of three layers trained using a support vector machine. Srinivas et al. developed a 7 layer fully convolutional neural network \cite{kruthiventi2017deepfix}. The network learns features in a pyramid form to predict saliency maps in an end-to-end pattern. Huang et al. expoited deep features of a pre-trained AlexNet, GoogleNet, and VGG-16 to train a saliency model \cite{huang2015salicon}. The model combines fine and coarse scales of the pre-trained features, then the model is trained using support vector machine. Jetley et al. formulated saliency maps as generalized berbulli distributions \cite{jetley2016end}. The architecture of the network is formulated by CNN with convolutional part of the layers identical to VGG. The model is trained using novel cost functions that compute the distances between probability distributions. Kummerer et al. developed a saliency model using AlexNet \cite{kummerer2014deep}. The model truncates the last three layers of the network and linearly combines all response images. Later, Kummerer et al. used the pre-trained deep features of VGG-19 to train a saliency model \cite{kummerer2016deepgaze}. The pre-trained features are fed to a 5 layers of $1\times1$ convolutional layers. The model is trained using a maximum likelihood learning scheme. Liu \& Han developed a saliency model by exploiting deep features of VGG or ResNet network as fine scale and placing CNN as coarse scale \cite{liu2016deep}. The two scales of features are fed to two long short term memory (LSTM) RNNs, and are trained using gradient descent. Cornia et al. extracted deep features from dilated VGG/ResNet, then fed the features to an LSTM recurrent network selectively attending different regions of a tensor without the concept of time \cite{cornia2016predicting}. Pan et al. proposed a saliency model as a convolutional encoder-decoder architecture \cite{pan2015end}. The encoder part of the model consists of VGG pre-trained features. The decoder part consists of upsampling followed by convolution filters. The model is trained by back-propagating the binary cross entropy as the cost function.

In this study, we explore the intuition of pre-trained deep feature without further training. In addition, we exploit the semantic information provided by fully connected layers to reflect the prior knowledge.

\subsection{Contributions of this study}
In this study, the contributions are threefold. First, a computational saliency model is proposed to predict human fixations using pre-trained deep features, codenamed DeepFeat. To our knowledge this is the only model that uses pre-trained deep features without further training. Second, four implementations of the DeepFeat are computed and compared to investigate the role of the pre-trained deep features in saliency prediction. Third, through extensive evaluation over four evaluation metrics and 9 state-of-the-art saliency models, we demonstrate that the DeepFeat model performs at the state-of-the-art level. 

\begin{figure*}
\includegraphics[width=\textwidth]{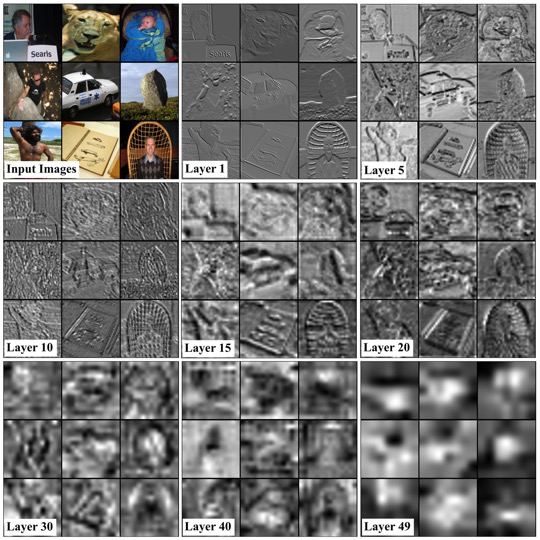}
\caption{Visualization of example features of layers 1, 10, 20, 30, 40, and 49 of a deep convolutional neural network. In each layer visualized, one convolution response image is selected randomly and presented.}
\end{figure*}

\section{Proposed Approach}

\begin{figure*}
\includegraphics[width=\textwidth,height=9cm]{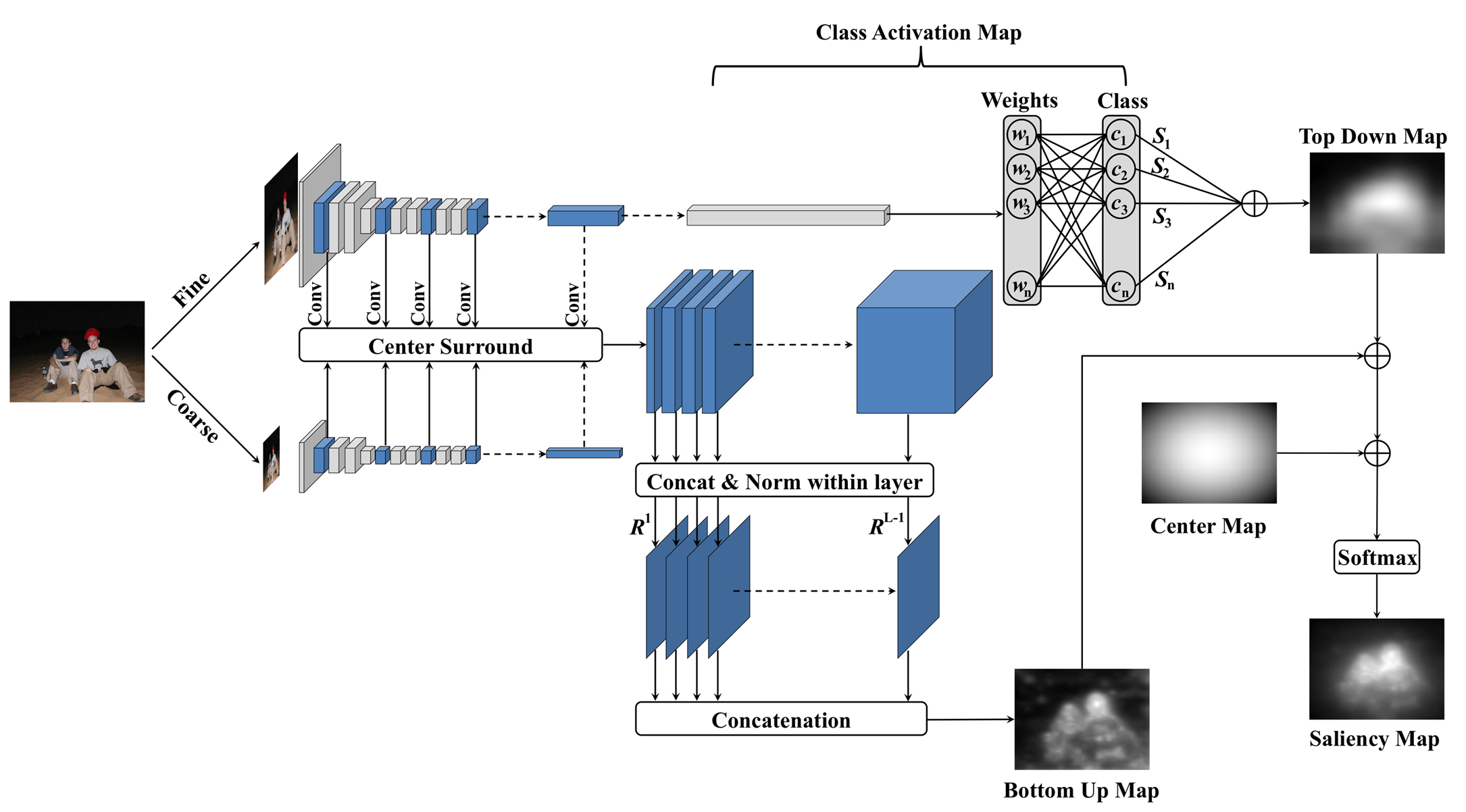}
\caption{Architecture of the DeepFeat Model. The architecture consists of a combination of bottom up and top down features. The bottom up features are computed using two scales of a CNN. The top down map is computed using the class activation map of the full scale CNN. The result of combining the bottom up and top down map is weighted by a center bias map.}
\end{figure*}

\subsection{Visualization of the deep features}
In this work, we obtained pre-trained deep features from a 50 layer residual network (ResNet) \cite{he2016deep}. The network was trained on ImageNet2012 dataset \cite{russakovsky2015imagenet} that consists of 1.28 million images of 1000 classes. The architecture of the network consists of a single convolution layer followed by batch normalization, rectified linear unit (ReLU), and max pooling layer. Then a series of residual shortcut building blocks. Visualization of the architecture can be found online \cite{resnet50}. In this study, residual shortcuts are avoided  as they add more complexity to the proposed bottom up architecture. Also, the response images of 49 convolution operations are used as deep features for the bottom up saliency computation. All computations were done in MatConvNet \cite{vedaldi2015matconvnet}. Fig. 1 presents example deep features of nine representative images from layers 1, 10, 20, 30, 40, and 49 of a residual network.

\subsection{DeepFeat architecture}
In this section, we formalize DeepFeat as a fusion of bottom up and top down visual factors using a simple combination strategy. The architucture of the DeepFeat can be visualized in Fig. 2.

A bottom up visual cues are represented by a CNN pre-trained features. For the purpose of bottom up computation, the fully connected layer of the CNN is removed. Previous studies suggest that the computation of two scales of CNN extracts semantic information about the image \cite{huang2015salicon, liu2016deep}. Therefore, two scales of the deep features are exploited, fine and coarse scales. The fine scale is original size of the extracted deep feature. The coarse scale is the downsampled version of the extracted deep feature. Inspired by Itti's work \cite{itti1998model}, the center-surround of the coarse and the fine scale for convolution response images is formed by:

\begin{equation}
R^\ell = \sum_{i=1}^{k}\abs{r_{0}^\ell(i) - r_{1}^\ell(i)}
\end{equation}

where $r_{0}$ denotes the fine scale feature, $r_{1}$ denotes the upsampled coarse scale feature, $i$ denotes the convolution response image of layer $\ell$, and $k$ denotes the number of response images in layer $\ell$. The total response $R$ at layer $\ell$ is normalized from 0 to 1, and then linearly combined by:

\begin{equation}
M_{Bottom Up} = \sum_{\ell=1}^{L-1}\mathcal{N}\bigg(R^\ell\bigg)
\end{equation}

where $L$ denotes the total number of layers in the network, and $\mathcal{N}(\cdot)$ is the normalization operator. In this study, the total response of convolutions is suggested to contribute equally in all bottom up layers.

In the top down map, the fully connected layer is exploited to emphasize the top down component of the network. The final output of a CNN is a softmax based probabilistic vector. Such vector represents the probability of classes for image recognition. The intuition of this work is to emphasize the top down component by extracting individual class. Following the class activation map (CAM) \cite{zhou2016learning}, the response images of the final activation in the network are multiplied by weights of the fully connected convolution filter:

\begin{equation}
CAM_{c}(x,y) = \sum_{k}w_{k,c}^La_{k}^{L-1}(x,y)
\end{equation}

where $c$ denotes a class of objects, $k$ denotes the number of units in the activation $a$ and the weight $w$, $x$ and $y$ denote the spatial location. The CAM detects a specific class in the image. In order to project all available classes on the image, the CAM of a class is weighted by its corresponding probability at the final fully connected layer:

\begin{equation}
M_{Top Down} = \sum_{c=1}^{C}S_{c}CAM_{c}
\end{equation}

where $C$ denotes the total number of classes, and $S$ denotes the softmax probability of the classes at the final fully connected layer.

The bottom up and top down maps are linearly combined:

\begin{equation}
Y = (1-\alpha)M_{Top Down} + \alpha M_{Bottom Up}
\end{equation}

where $\alpha$ is a constant equal to $0.5$ in this study. To account for human bias toward the center in visual strategies \cite{parkhurst2003scene, tatler2005visual, tatler2007central}, a center bias is incorporated by:

\begin{equation}
Y' = (1-\beta)Y + \beta M_{center}
\end{equation}

where $\beta$ is a constant equal to $0.5$ in this study, and $M_{center}$ is a center bias map computed using a Gaussian kernel with a cut off frequency equivalent to the maximum dimension of the image. Finally, a probability distribution of the saliency map is obtained using softmax:

\begin{equation}
S = \frac{e^{Y'(x,y)}}{\sum_{x,y}e^{Y'(x,y)}}
\end{equation}

\begin{figure*}
\includegraphics[width=\textwidth]{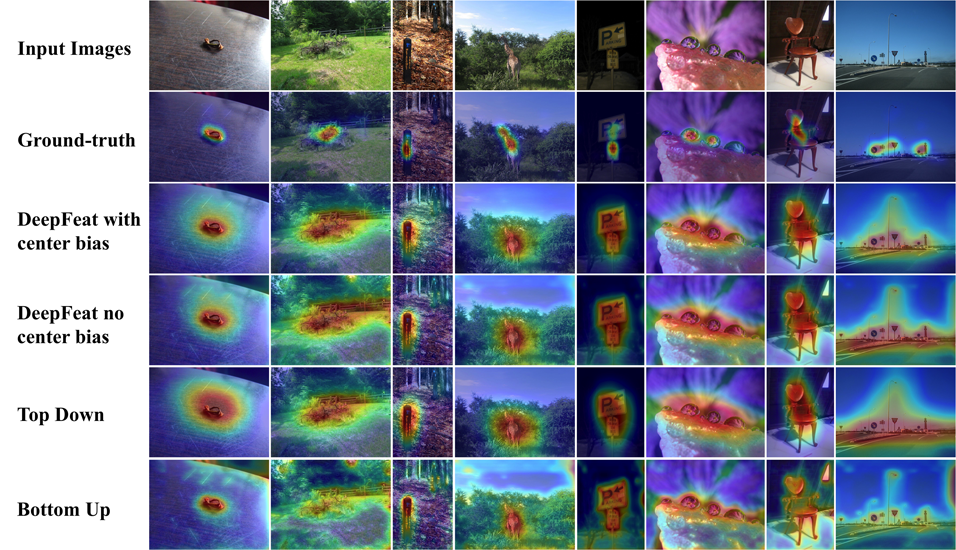}
\caption{Row 1 show photographs of input images. Row 2 show the corresponding ground-truth fixation maps. Row 3 to 6 show four saliency maps obtained by the DeepFeat model.}
\end{figure*}

\section{Experimental Results}

\subsection{Experimental setup}

\subsubsection{Dataset}
In this study, we used the MIT1003 dataset to validate the proposed approach. The MIT1003 dataset consists of 1003 images. The resolution of the images is fixed on one dimension 1024 pixels, and on the other dimension it ranges from 678 to 768. Fifteen observers freely viewed the MIT1003 images. Images are presented to 15 observers for 3 seconds \cite{judd2009learning}.

\subsubsection{Evaluation metrics}
Saliency models are usually evaluated by comparing their predictions to human fixation maps using evaluation metrics. In this work, the proposed framework is evaluated using four evaluation metrics \cite{bylinskii2016different}.

\subsubsection*{ROC}
is a binary classification measure of the intersected area between the predicted saliency and human fixations. At various thresholds, the trade-off between the true and false positive rates is plotted. The true and false positives (TPR and FPR, respectively) are formed by:

\begin{equation}
TPR = \frac{TP}{TP+FN}
\end{equation}

\begin{equation}
FPR = \frac{FP}{FP+TN}
\end{equation}

Two ROC variants are exploited to measure the intersection between the saliency and human fixations. The first ROC variant measures the intersection between a saliency map and a ground-truth distribution of human fixation. In the second variant, a uniform random sample of image pixels is used as negatives. Then, one defines the saliency map values above threshold at these pixels as false positives.

\subsubsection*{AUC}
is an integration of the spatial area under the ROC curve. Such that, the random guessing score is 0.5. A score above 0.5 indicates the predictions are above random guessing. Two variants of AUC are presented in this study. The first variant (AUC), is the integral of the first ROC variant, and the second AUC variant (AUC Borji), is the integral of the second ROC variant.

\subsubsection*{CC}
is a measure of the statistical relationship between the predicted saliency map and the human ground-truth. The saliency map and human ground-truth are treated as random variables, and the strength and direction between the two variables are measured by: 

\begin{equation}
CC(S,F)=\frac{cov(S,F)}{\sigma(S)\sigma(F)}
\end{equation}

where $cov(S,F)$ denotes the covariance between the saliency map $S$ and the human ground-truth $F$. A score of -1 or 1 indicates a perfect correlation between the two maps. A score of zero indicates the two maps are not correlated.

\subsubsection*{KL} is a measure of how the predicted saliency map diverge from the human ground-truth map in a probabilistic interpretation of the two maps:

\begin{equation}
KL(S,F)=\displaystyle\sum F log(\epsilon+\frac{F_{i}}{\epsilon+S})
\end{equation}

where $\epsilon$ is a constant. A score of 0 indicates the two maps are identical. A positive score indicates the divergence between the two maps.

\begin{table*}
\centering
\captionsetup{width=11cm}
\caption{Ranking of four implementations of the DeepFeat model over four evaluation metrics.}
\label{my-label}
\begin{tabular}{lcccc}
\hline
               & AUC               & AUC Borji         & CC                & KL                \\ \hline
DeepFeat (WCB) & 0.857 $\pm$ 0.002 & 0.835 $\pm$ 0.002 & 0.443 $\pm$ 0.004 & 1.412 $\pm$ 0.009 \\
DeepFeat(NCB)  & 0.782 $\pm$ 0.004 & 0.777 $\pm$ 0.004 & 0.338 $\pm$ 0.005 & 1.550 $\pm$ 0.01  \\
DeepFeat (TD)  & 0.715 $\pm$ 0.004 & 0.751 $\pm$ 0.004 & 0.309 $\pm$ 0.006 & 1.555 $\pm$ 0.012 \\
DeepFeat (BU)  & 0.776 $\pm$ 0.004 & 0.743 $\pm$ 0.004 & 0.283 $\pm$ 0.005 & 1.652 $\pm$ 0.009 \\ \hline
\end{tabular}
\end{table*}

\begin{table*}
\centering
\caption{Ranking of DeepFeat model and 9 saliency models.}
\label{my-label}
\begin{tabular}{lcccc}
\hline
         & AUC               & AUC Borji         & CC                & KL                \\ \hline
DeepFeat & 0.857 $\pm$ 0.002 & 0.834 $\pm$ 0.002 & 0.443 $\pm$ 0.004 & 1.412 $\pm$ 0.009 \\
AWS      & 0.712 $\pm$ 0.004 & 0.743 $\pm$ 0.004 & 0.322 $\pm$ 0.007 & 1.54 $\pm$ 0.0153 \\
BMS      & 0.747 $\pm$ 0.003 & 0.768 $\pm$ 0.004 & 0.357 $\pm$ 0.006 & 1.452 $\pm$ 0.012 \\
CovSal   & 0.736 $\pm$ 0.002 & 0.752 $\pm$ 0.003 & 0.408 $\pm$ 0.006 & 1.622 $\pm$ 0.026 \\
eDN      & 0.863 $\pm$ 0.002 & 0.845 $\pm$ 0.002 & 0.41 $\pm$ 0.003  & 1.545 $\pm$ 0.01  \\
GBVS     & 0.827 $\pm$ 0.002 & 0.813 $\pm$ 0.003 & 0.417 $\pm$ 0.005 & 1.297 $\pm$ 0.01  \\
Judd     & 0.843 $\pm$ 0.002 & 0.830 $\pm$ 0.002 & 0.417 $\pm$ 0.003 & 1.547 $\pm$ 0.009 \\
ML-Net   & 0.668 $\pm$ 0.003 & 0.772 $\pm$ 0.003 & 0.592 $\pm$ 0.007 & 1.344 $\pm$ 0.027 \\
RARE     & 0.747 $\pm$ 0.003 & 0.771 $\pm$ 0.004 & 0.379 $\pm$ 0.006 & 1.415 $\pm$ 0.014 \\
UHF      & 0.821 $\pm$ 0.003 & 0.811 $\pm$ 0.003 & 0.416 $\pm$ 0.005 & 1.407 $\pm$ 0.011
\\
\hline
\end{tabular}
\end{table*}

\subsection{Analysis of the architecture}
Fig. 3 presents four implementations of the proposed DeepFeat model. The implementations are: bottom up (BU), top down (TD), a saliency map without center bias (NCB), and a saliency map with center bias (WCB). To quantitatively analyze the four saliency maps, we draw AUC, AUC borji, CC, and KL scores over the MIT1003 dataset in Fig. 4. To measure the statistical significance of mean scores between two consecutive models, a t-test is used at a the significance rate of $p \leq 0.5$.

In Fig. 4, the ranking of the four implementations is consistent across all scores. The WCB implementation significantly outperforms the other implementations of the DeepFeat models. It highlights that the importance of adding a center bias to weight the prediction of human fixations. The NCB implementation slightly outperforms the TD implementation over AUC, AUC Borji, and KL scores, and significantly outperforms the TD implementation over the CC score. Such results occur because the NCB implementation emphasizes intersected locations between the TD and BU implementations. The TD implementation significantly outperformed the BU saliency maps over all four scores. This may occur because majority of the human fixations are explained by top down factors rather than bottom up. The complete description of the comparison is provided in Table 1.

Generally speaking, not only the WCB implementation scores the highest across all four implementations of the DeepFeat model, but also achieves the smallest margin of error. The standard error of the mean for the WCB implementation is the smallest across all implementations of the DeepFeat model over all four evaluation metrics. Such result confirms the human fixations tends to have intense bias toward the center of an image.

\begin{figure}
\includegraphics[width=8.5cm,height=7.5cm]{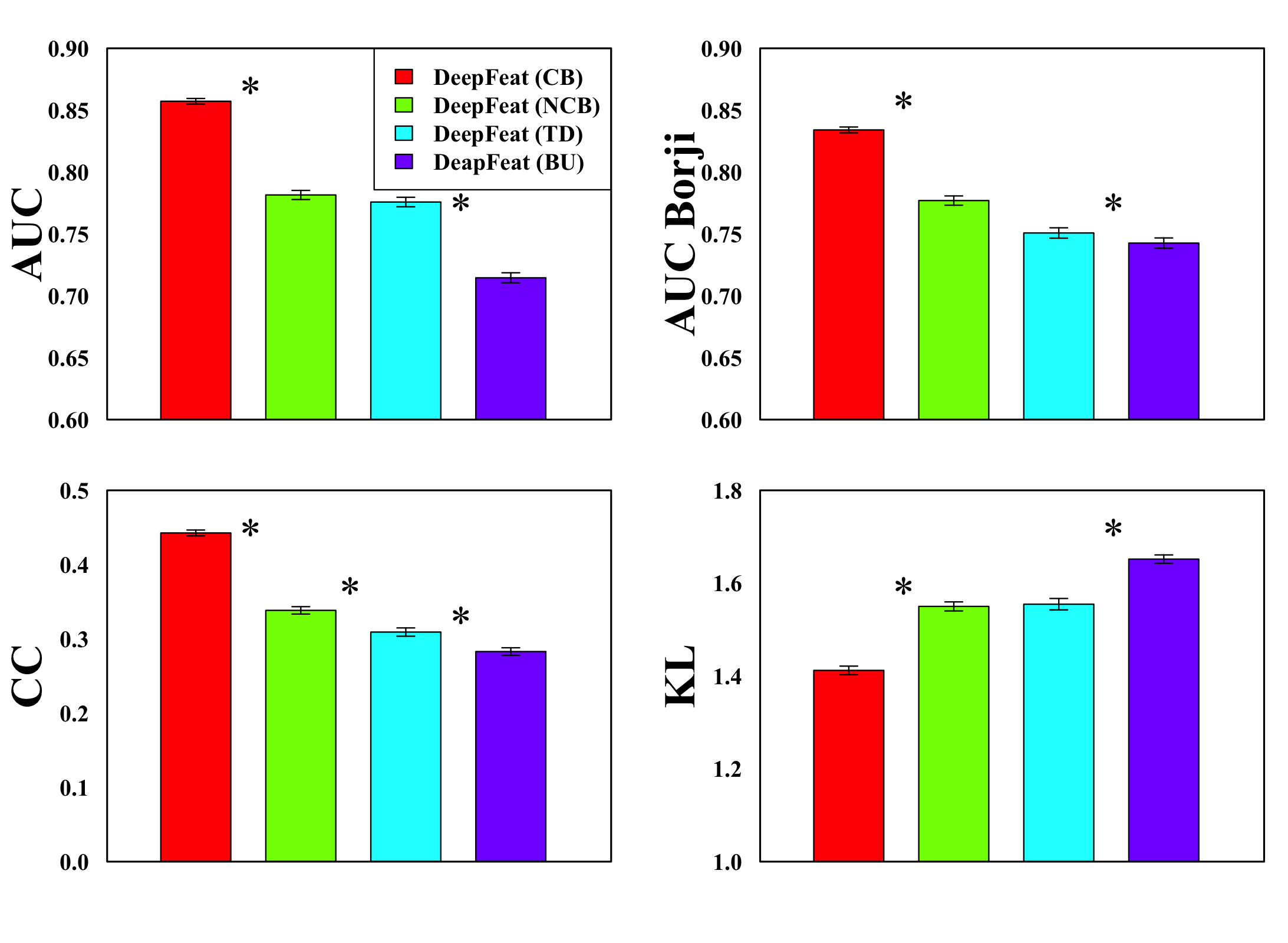}
\caption{Averaged scores of four implementations of the proposed DeepFeat model using four evaluation metrics: AUC, AUC Borji, CC, and KL. A * indicates the two consecutive models are significantly different using t-test at confidence level of $p \leq 0.05$. Models that are not consecutive have a larger probability to achieve statistical significance. Standard error of the mean (SEM) is indicated by the error bars.}
\end{figure}

\begin{figure*}
\includegraphics[width=\textwidth]{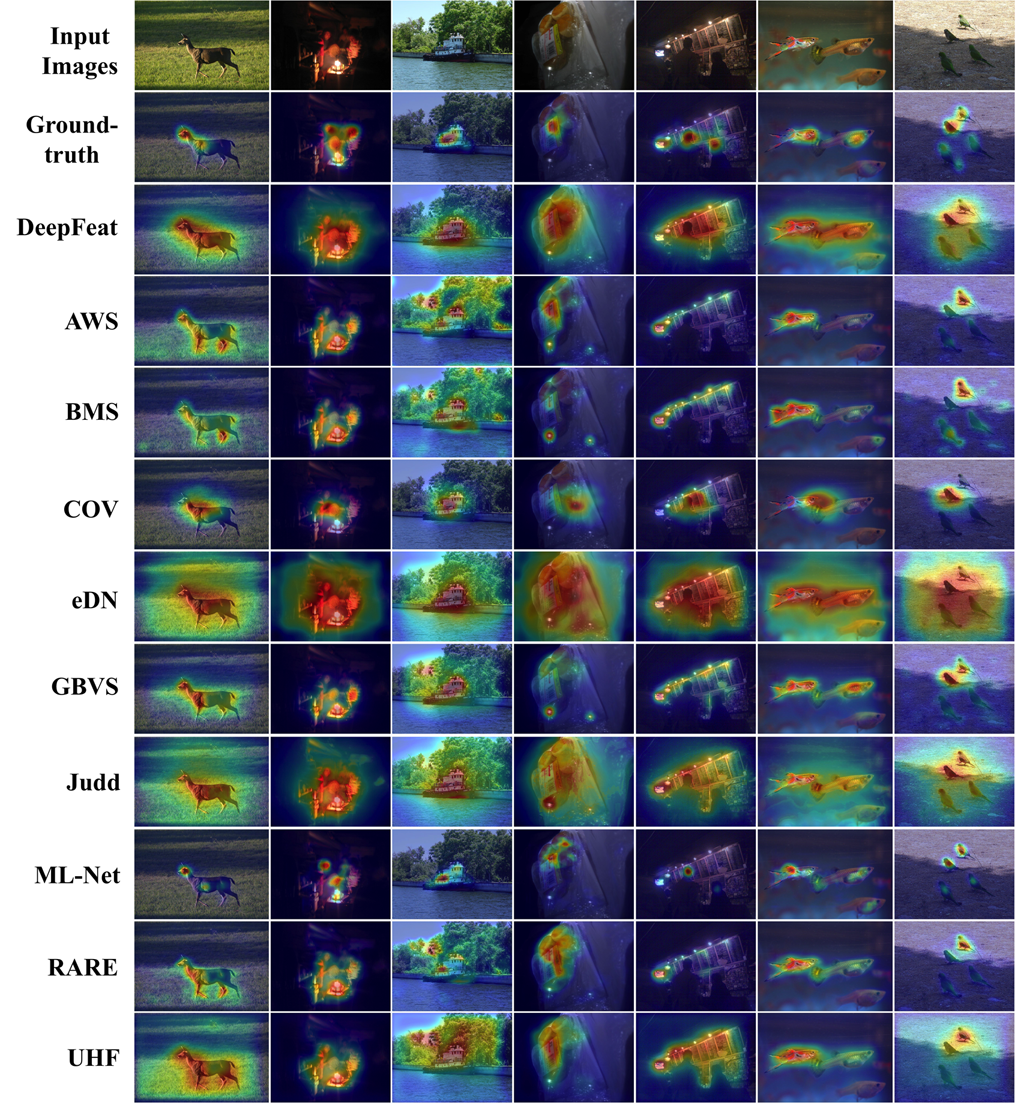}
\caption{Row 1 show the photographs of ten input images in MIT1003 dataset. Row 2 show the corresponding ground-truth fixation maps. Saliency maps computed by the proposed DeepFeat model are shown in row 3. Rows 4 to 12 present saliency maps computed by 9 other state-of-the-art saliency models.}
\end{figure*}

\subsection{Comparison with other state-of-the-art saliency models}
In this section, the DeepFeat model is compared to nine state-of-the-art saliency models including four learning based saliency models eDN \cite{vig2014large}, Judd \cite{judd2009learning}, ML-Net \cite{cornia2016deep}, and UHF \cite{tavakoli2016bottom}, and five classical saliency models AWS \cite{garcia2012relationship}, BMS \cite{zhang2016exploiting}, CovSal \cite{erdem2013visual}, GBVS \cite{harel2007graph}, and RARE \cite{riche2013rare2012}. Fig. 5 illustrates sample images of the tested dataset along with the corresponding saliency maps of 10 saliency models (DeepFeat and the 9 saliency models). Fig. 6 presents the ROC curves and the AUC scores of DeepFeat and other 9 saliency models. In the top charts of Fig. 6, it is clear that the DeepFeat model outperforms most saliency models. To summarize and further investigate, we draw the AUC, and AUC Borji in the bottom charts of Fig. 6. For statistical significance test of mean scores, a t-test is used at $p \leq 0.5$ level of significance. Although the models ranking order is not identical for both charts, some general patterns can be observed. In both scores, the eDN ranks first and the DeepFeat ranks second. This may occur because eDN and DeepFeat incorporate center bias to their predictions. The top four models in both scores incorporate a center bias. Although the ML-Net model incorporates a learned center bias, it is ranked tenth on AUC score, and seventh on AUC Borji.

To further investigate the performance of the proposed DeepFeat model, we draw the CC and KL scores for the 10 saliency models in Fig. 7. Using the CC score, ML-net is ranked first, and DeepFeat is ranked second. Compared with the CC score of the DeepFeat model, a significantly larger CC score of the ML-Net model occurs because the ML-Net model incorporates a learned center bias map. Such map has a larger correlation with the human eye fixation than a 2D Gaussian distribution (center bias map of DeepFeat). Using the KL score, GBVS ranks first, ML-Net ranks second, UHF ranks third, and DeepFeat ranks fourth. This is because the prediction region of the DeepFeat model is large. The large area of prediction occurs in the top down map while predicting objects of the image.

The overall performance indicates that the proposed DeepFeat model is among the highest ranking saliency models over the four scores. In addition to the prediction scores, the DeepFeat model takes 150s to predict one saliency map with CPU (Core i7 2.3GHz and 8GB RAM).

\begin{figure}
\includegraphics[width=9cm,height=7.5cm]{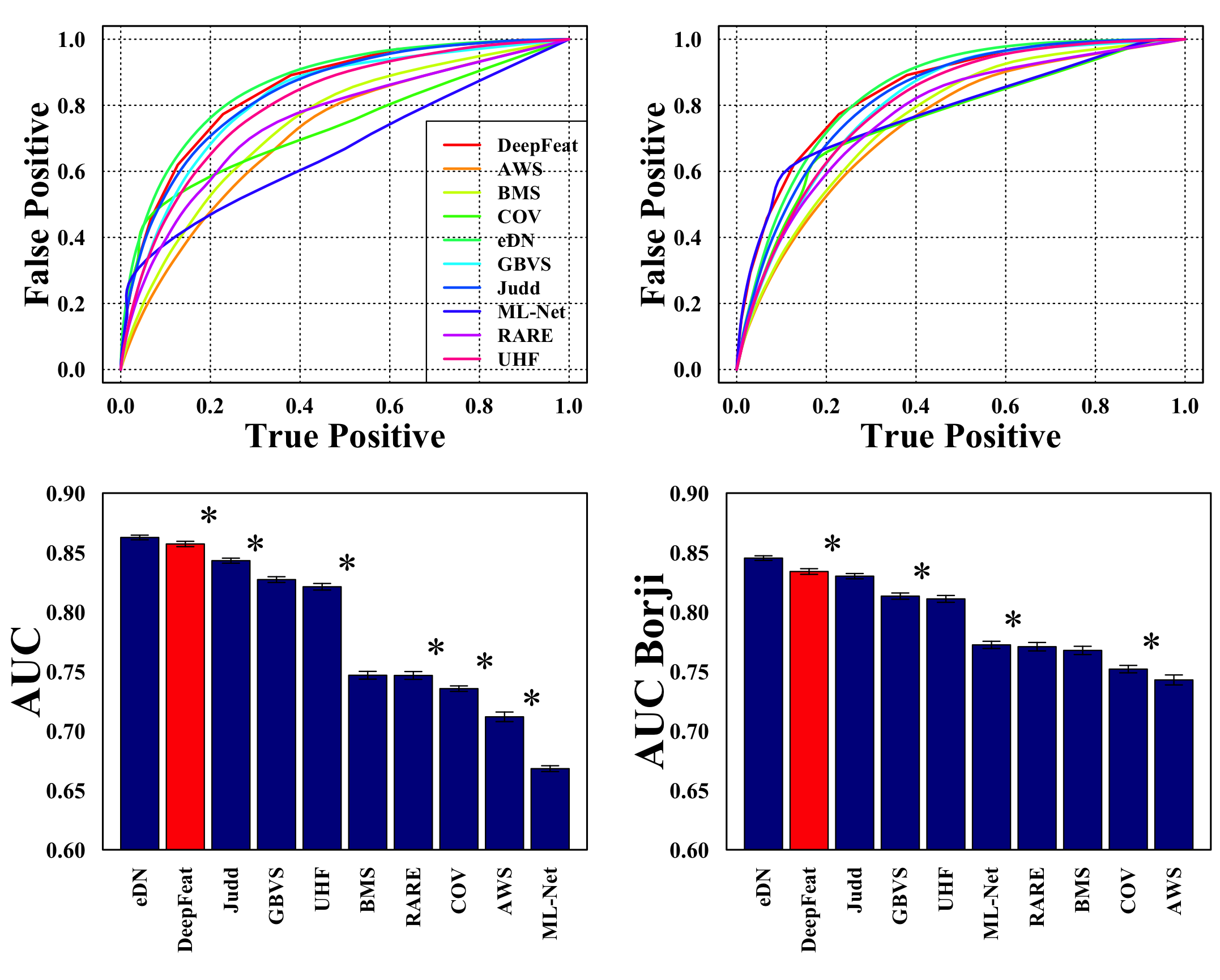}
\caption{Averaged ROC curves (top charts) and AUC values (bottom charts) of ten saliency models including the proposed DeepFeat model and 9 other saliency models. Two variants of the ROC and AUC are included. The first variant (right charts) is based on a distribution based ground-truth map, and the second variant (left charts) is based on a fixation points ground-truth map. The bottom charts indicate that the ranking of ten saliency maps over  MIT1003 dataset using AUC and AUC Borji. A * indicates the two consecutive models are significantly different using t-test at confidence level of $p \leq 0.05$. Models that are not consecutive have a larger probability to achieve statistical significance. SEM is indicated by the error bars.}
\end{figure}

\begin{figure}
\includegraphics[width=9cm,height=3.5cm]{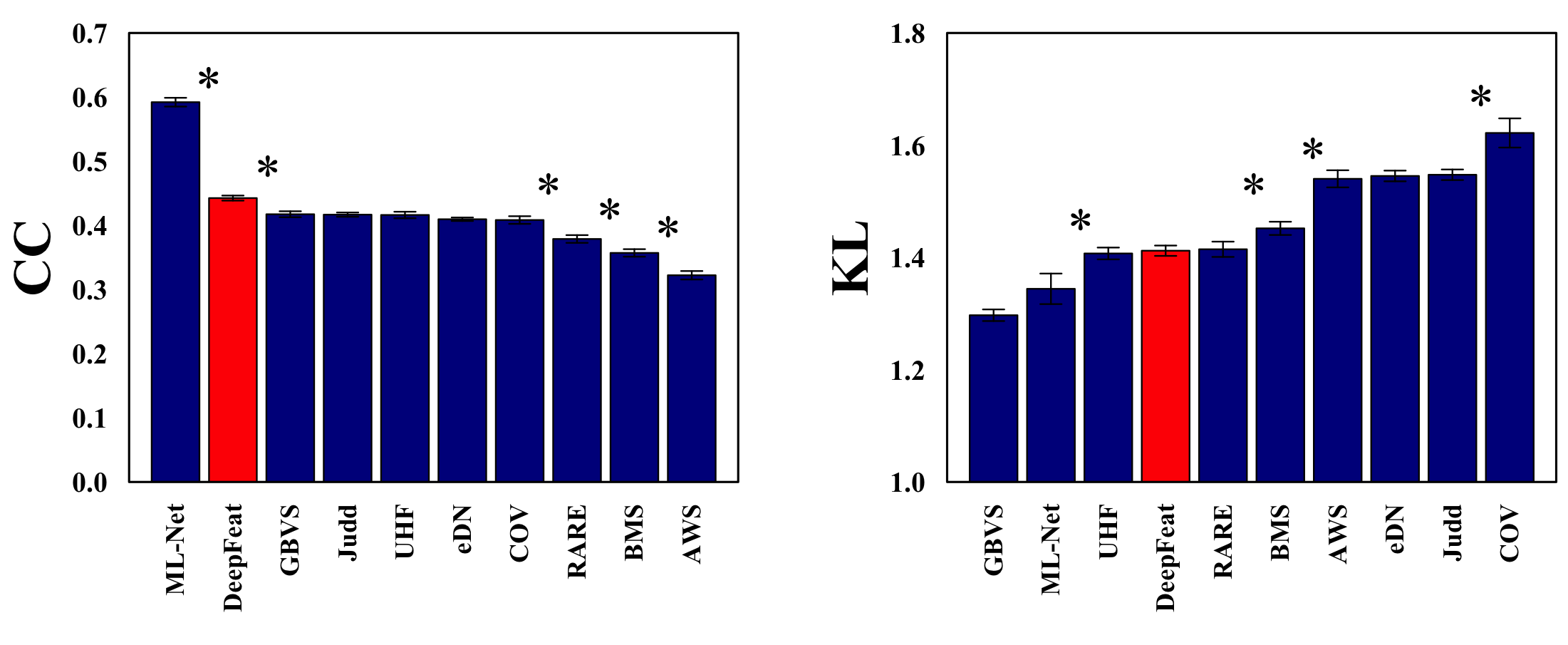}
\caption{Ranking of the DeepFeat model and nine other saliency models over MIT1003 dataset using CC and KL scores. A * indicates the two consecutive models are significantly different using t-test at confidence level of $p \leq 0.05$. Models that are not consecutive have a larger probability to achieve statistical significance. SEM is indicated by the error bars}
\end{figure}

\section{Conclusion}
In this study, we proposed a deep feature based saliency model, codenamed DeepFeat, which combines bottom up and top down visual factors obtained from pre-trained deep features. To validate the performance of the DeepFeat model, we investigated four different implementations of the DeepFeat model using four evaluation metrics over the MIT1003 dataset. The results demonstrate that the implementation of the DeepFeat model with incorporation of center bias outperforms all other three implementations. Moreover, we also evaluated performance of the proposed DeepFeat model compared with 9 other state-of-the-art saliency models using four evaluation metrics over the MIT1003 dataset. The experimental results show that the proposed DeepFeat model ranks among the top saliency models. In future work, we will examine more popular CNNs such as VGG and GoogLeNet. Also, response images from activation maps, pooling, batch normalization, etc., will be validated. Moreover, the performance of the DeepFeat will be evaluated with other datasets.

\ifCLASSOPTIONcaptionsoff
  \newpage
\fi

\bibliographystyle{IEEEtran}
\bibliography{references}

\end{document}